\documentclass{article}

\usepackage{PRIMEarxiv}

\usepackage[utf8]{inputenc} 
\usepackage[T1]{fontenc}    
\usepackage{hyperref}       
\usepackage{url}            
\usepackage{booktabs}       
\usepackage{amsfonts}       
\usepackage{nicefrac}       
\usepackage{microtype}      
\usepackage{lipsum}
\usepackage{fancyhdr}       
\usepackage{graphicx}       
\graphicspath{{media/}}     

\usepackage{amsmath,amsfonts}
\usepackage{algorithm}
\usepackage{array}
\usepackage[caption=false,font=normalsize,labelfont=sf,textfont=sf]{subfig}
\usepackage{textcomp}
\usepackage{stfloats}
\usepackage{url}
\usepackage{verbatim}
\usepackage{graphicx}
\usepackage{cite}
\usepackage[dvipsnames]{xcolor}

\usepackage{amssymb, mathtools, amsthm, xspace, multirow, makecell, bbm, booktabs, algorithmicx}
\usepackage[group-separator={,}, group-minimum-digits={3}]{siunitx}
\usepackage[noend]{algpseudocode}
\makeatletter
\DeclareRobustCommand\onedot{\futurelet\@let@token\@onedot}
\def\@onedot{\ifx\@let@token.\else.\null\fi\xspace}

\def\eg{\emph{e.g}\onedot}

\def\etc{\emph{etc}\onedot}

\def\etal{\emph{et al}\onedot}
\makeatother

\newcommand{\norm}[1]{\left\lVert#1\right\rVert}
\DeclareMathOperator{\E}{\mathbb{E}}

\newcommand{\INPUT}{\textbf{Input:}} 
\newcommand{\OUTPUT}{\textbf{Output:}} 

\title{DiFace: Cross-Modal Face Recognition through Controlled Diffusion
\thanks{This work has been submitted to the IEEE for possible publication. Copyright may be transferred without notice, after which this version may no longer be accessible.} 
}

\author{
  Bowen~Sun, Shibao~Zheng \\
  Department of Electronic Engineering of SEIEE \\
  Shanghai Jiao Tong University \\
  Shanghai, China\\
  \texttt{\{sunbowen, sbzh\}@sjtu.edu.cn} \\
}

\begin{document}
\maketitle

\begin{abstract}
Diffusion probabilistic models (DPMs) have exhibited exceptional proficiency in generating visual media of outstanding quality and realism. 
Nonetheless, their potential in non-generative domains, such as face recognition, has yet to be thoroughly investigated. 
Meanwhile, despite the extensive development of multi-modal face recognition methods, their emphasis has predominantly centered on visual modalities. 
In this context, face recognition through textual description presents a unique and promising solution that not only transcends the limitations from application scenarios but also expands the potential for research in the field of cross-modal face recognition. 
It is regrettable that this avenue remains unexplored and underutilized, a consequence from the challenges mainly associated with three aspects: 1) the intrinsic imprecision of verbal descriptions; 2) the significant gaps between texts and images; and 3) the immense hurdle posed by insufficient databases.
To tackle this problem, we present DiFace, a solution that effectively achieves face recognition via text through a controllable diffusion process, by establishing its theoretical connection with probability transport.
Our approach not only unleashes the potential of DPMs across a broader spectrum of tasks but also achieves, to the best of our knowledge, a significant accuracy in text-to-image face recognition for the first time, as demonstrated by our experiments on verification and identification.
\end{abstract}
\keywords{Cross-modal, face recognition, diffusion probabilistic models}
\section{Introduction}
\label{sec:intro}
In contemporary artificial intelligence (AI), generative models~\cite{goodfellowGenerativeAdversarialNetworks2020a, sohl-dicksteinDeepUnsupervisedLearning2015} and multi-modal learning emerge as thriving domains.
As a prominent and blooming field within generative AI, DPMs, also referred to as diffusion models, have exhibited exceptional prowess in the realm of content generation, effectively generating visually stunning and realistic media of superior quality.
Noteworthy contributions, \eg, image generation~\cite{hoDenoisingDiffusionProbabilistic2020, songScoreBasedGenerativeModeling2021}, audio synthesis~\cite{chenWaveGradEstimatingGradients2020}, video generation~\cite{hoImagenVideoHigh2022} and data purification~\cite{nieDiffusionModelsAdversarial2022a}, have solidified their presence in various fields that require the application of generative artificial intelligence.
Multi-modal content analysis~\cite{ouMultimodalLocalGlobalAttention2021, zhuMultimodalDeepAnalysis2020} and generation~\cite{marivaniDesigningCNNsMultimodal2022}, have further garnered significant attention, in consideration of the diverse modalities from which human cognition originates.
The advent of text-to-image models endowed with controllable generation, exemplified by Stable Diffusion (SD)~\cite{rombachHighResolutionImageSynthesis2022a} and DALL-E~\cite{rameshZeroShotTexttoImageGeneration2021, rameshHierarchicalTextconditionalImage2022}, has revolutionized the multi-modal generation, ushering in newfound abilities for creative endeavors. 
By leveraging the power of DPMs, these notable achievements expand the boundaries of artistic creation and have the possibility to enhance assorted industries.

\begin{figure}[t]
    \centering
    \includegraphics[width=\linewidth]{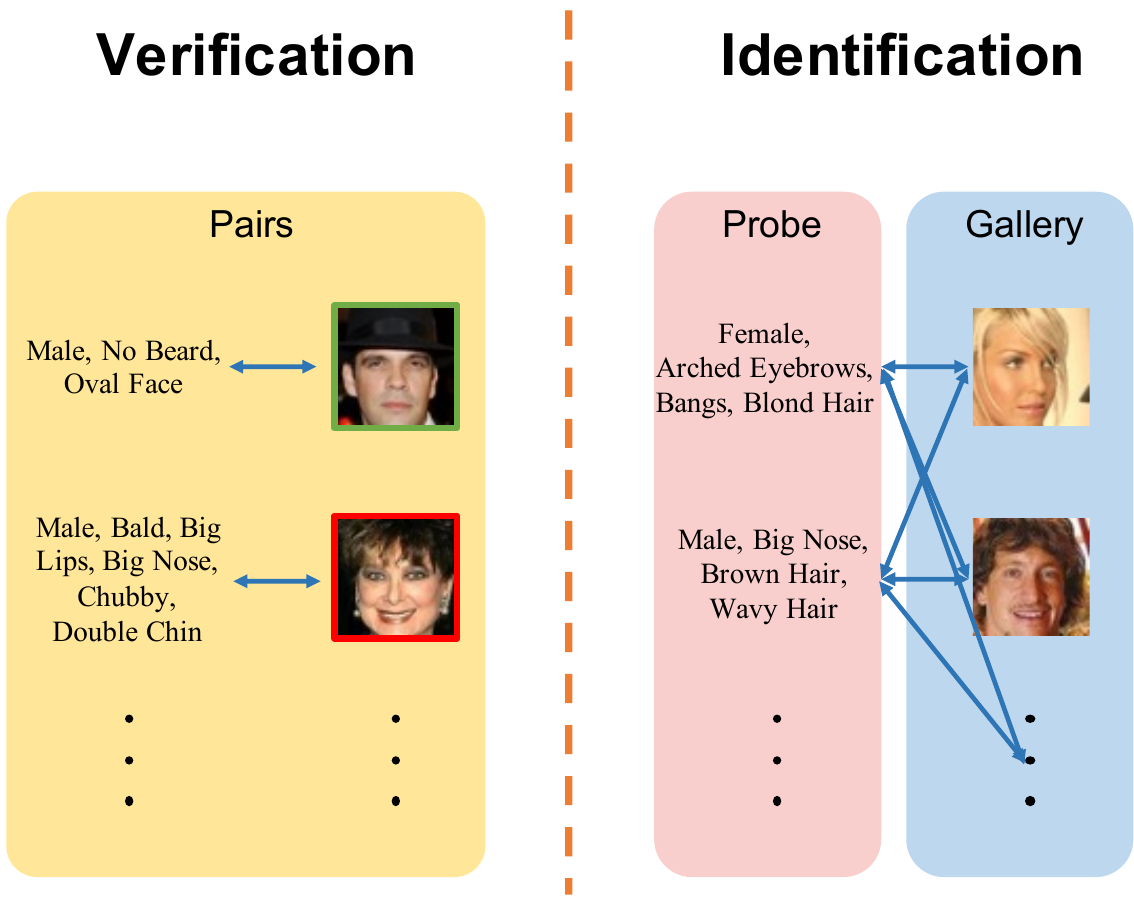}
    \caption{The illustration of text-to-image face recognition including verification (left) and identification (right). During the verification process, the model assesses whether each pair of textual description and facial image pertains to the same subject (green frame) or different subjects (red frame). In the identification phase, the recognition model compares each verbal description in the probe set with all the images in the gallery to rank the corresponding similarity scores.}
    \label{fig:illustration}
\end{figure}

While diffusion models excel at capturing the intricate details for synthesis, their potential in extensive domains irrelevant to generation, such as face recognition, is yet to be fully explored. 
Traditional face recognition methods relying on normal RGB images~\cite{dengArcFaceAdditiveAngular2019, wangCosFaceLargeMargin2018} have achieved high accuracy with limited scope for further enhancement though, the complication of cross-modal recognition~\cite{mingCrossmodalPhotocaricatureFace2021,koleyCrossmodalFaceRecognition2023} pose a significant bottleneck that is widely acknowledged and considered essential in advancing the field.
One intriguing approach to cross-modal face recognition is face recognition by textual descriptions illustrated in Fig.~\ref{fig:illustration}, which holds immense value in numerous scenarios, spanning from public security applications to object retrieval.
It becomes feasible to establish a connection between visual and textual modalities, facilitating identity filtering solely based on verbal descriptions. 
This capability effectively resolves an otherwise insurmountable difficulty arising from the absence of visual information.

Regrettably, existing applications of diffusion models are completely reliant on their generative capability and cross-modal face recognition predominantly encompasses visual information across diverse modalities.
Contemporary text-guided generative models typically employ language modules absorbing prompts~\cite{radfordLearningTransferableVisual2021} or natural languages~\cite{brownLanguageModelsAre2020} to exert influence on the diffusion directions, thereby enabling the creation of vibrant images through pre-trained generation modules based on variational autoencoders (VAEs)~\cite{kingmaAutoEncodingVariationalBayes2022} or generative adversarial networks (GANs)~\cite{goodfellowGenerativeAdversarialNetworks2020a}.
VAEs and GANs have emerged as effective frameworks for learning rich latent representations and generating high-quality images thus serve as a crucial role in shaping the overall outcomes.
During the intermediate process of the controllable diffusion, initial random noises are skillfully conveyed and channeled into the latent space of generation modules by encoded word embeddings that capture the semantic meaning of the given text. 
On the other hand, current multi-modal face recognition primarily focuses on various aspects, including near-infrared~\cite{heLearningInvariantDeep2017, heWassersteinCNNLearning2018}, forensic sketches~\cite{mittalCompositeSketchRecognition2015, galeaForensicFacePhotosketch2017}, depth imagery~\cite{kimDeep3DFace2017a, gilaniLearningMillions3D2018}, and caricature~\cite{mingCrossmodalPhotocaricatureFace2021}, \etc. 
These different aspects of multi-modal face recognition address the demands within their respective domains to a certain extent, contributing to the development of robust and versatile recognition systems capable of handling diverse modalities and real-world challenges.

Notwithstanding the imperative requirement for text-to-image face recognition, the enduring challenge of resolving this predicament remains unresolved, primarily due to the intricate complexities inherent in the task and the inadequacy of available data. 
The primary point is that verbal descriptions inherently lack the precision and richness of visual information, rendering cross-modal text-to-image recognition itself incapable of achieving the level of effectiveness achieved by direct image-to-image algorithms.
Additionally, in comparison to the relatively limited disparity observed among modalities within images, \eg, sketch-photo pairs, the divergence between textual and visual signals is considerably more substantial. 
This pronounced dissimilarity poses formidable obstacles in devising powerful recognition algorithms capable of effectively bridging the gap between textual and visual representations. 
Moreover, the scarcity of facial datasets that contain both comprehensive identity information and accompanying textual descriptions constitutes a formidable impediment to the advancement of related research endeavors.
The dearth of such databases, which simultaneously capture and integrate textual and visual information, significantly hampers the training and evaluation of models, thereby impeding the exploration of novel approaches and innovative solutions in this specialized domain.

In response to the growing demand for text-oriented face recognition, we propose the method named DiFace, which unleash the untapped potential of current diffusion models far limited by generation-centric employment.
We commence by presenting a probabilistic density movement as an elucidation of the mechanisms of diffusion models, deviating from the well-known Evidence Lower Bound (ELBO) viewpoint~\cite{luoUnderstandingDiffusionModels2022}. 
In this alternative perspective, we employ a theory of distribution transport to comprehend the fundamental mechanisms governing diffusion models.
By harnessing the power of this understanding, we have successfully achieved text-to-image face recognition without the need for any intermediate generation procedures, which allows us to instead directly utilize the capabilities of DPMs to accomplish the expected task. 
In order to augment the recognition capability of DPMs, we have additionally devised an additional refinement module, leading to the attainment of a final accuracy level of approximately $80\%$.
Rigorous and impartial experiments, encompassing verification and identification as benchmarks, have been meticulously conducted to showcase the effectiveness of DiFace. 
These findings not only demonstrate the possibility for DPMs to perform recognition tasks but also lay the foundation for future advancements in this particular domain.
The contributions can be summarized as follow:
\begin{itemize}
    \item We have achieved a noteworthy advancement in the field of cross-modal face recognition through the textual descriptions, a previously unexplored perspective. 
    \item Our approach creatively designs a refinement module, enabling the realization of recognition tasks via the probabilistic diffusion process, which circumvents the typical dependence on image synthesis.
    \item We offer a theoretical analysis as the cornerstone of this endeavor, establishing a vital linkage between probability diffusion flow and feature-based recognition. This expanded scope of application for generation-oriented DPMs emphasizes their substantial potential across broader domains.    
\end{itemize}


\section{Related Work}
\label{sec:related_work}

We review typical diffusion models and cross-modal face recognition methods in this section.

\subsection{Diffusion Models}
Drawing inspiration from the principles of nonequilibrium thermodynamics in physics, Sohl-Dickstein~\etal~\cite{sohl-dicksteinDeepUnsupervisedLearning2015} pioneer a generative model, serving as a precursor to subsequent DPMs, that tractably samples intricate data from simple distributions instead of earlier GAN~\cite{goodfellowGenerativeAdversarialNetworks2020a} algorithm.
In order to effectively synthesize high-quality images, Denoising Diffusion Probabilistic Models (DDPM)~\cite{hoDenoisingDiffusionProbabilistic2020} and Denoising Diffusion Implicit Models (DDIM)~\cite{songDenoisingDiffusionImplicit2022} facilitate the learning of neural networks from parameterized Markov chains. 
These chains are designed to reverse the diffusion process by adding noise to the data in the opposite direction of sampling until the signal is eliminated. 
Particularly, when this process gradually involves small amounts of Gaussian noise, it becomes feasible to set the transitions in the sampling chain as conditional Gaussian distributions.
A series of methods is subsequently proposed to enhance the efficiency of generation. 
Song~\etal~\cite{songGenerativeModelingEstimating2019, songScoreBasedGenerativeModeling2021} provide a score matching perspective to reformulate the previous Markovian process into a Stochastic Differential Equation (SDE), which in turn derives an Ordinary Differential Equation (ODE) using the Fokker-Planck equation (Kolmogorov's forward equation).
This viewpoint fosters the development of solvers~\cite{luDPMSolverFastODE2022} aimed at minimizing computational overhead and accommodates diverse ODE forms~\cite{liuGenPhysPhysicalProcesses2023, xuPFGMUnlockingPotential2023, xuPoissonFlowGenerative2022} that sample images from fundamental distributions.

With the advancements in natural language processing (NLP) propelled by the transformer~\cite{vaswaniAttentionAllYou2017}, neural networks have gained the ability to rapidly understand and generate contextually relevant conversations, thereby ushering in a new era for text-guided generation.
SD~\cite{rombachHighResolutionImageSynthesis2022a}, as one of latent diffusion models (LDMs), introduces a text-to-image generative technique that demonstrates strong scalability in producing highly detailed and efficient image synthesis. 
This multi-modal generation is achieved by compressing the higher-dimensional distribution of images into a lower-dimensional latent space accepted by an encoder/decoder~\cite{esserTamingTransformersHighResolution2021, yuVectorquantizedImageModeling2022} and employment of a diffusion process guided by word embeddings tokenized from the CLIP model~\cite{radfordLearningTransferableVisual2021}.
The emergence of similar techniques such as DALL-E~\cite{rameshZeroShotTexttoImageGeneration2021, rameshHierarchicalTextconditionalImage2022} enhances the flourishing of this domain.

The UNet, recognized as the prevailing architectural framework utilized in contemporary DPMs, was originally conceived with the specific objective of biomedical image segmentation~\cite{ronnebergerUNetConvolutionalNetworks2015}. 
Notably, this network is engineered to produce output that aligns precisely with the dimensions of the input, ensuring consistency and preserving the spatial information inherent in the probability distribution.
It has been adapted and enhanced for deployment in SD, where token-based conditioning mechanisms are utilized to exert control over the diffusion process. 
The UNet structure, with the flexible tokenizer, enables the incorporation of these conditioning mechanisms, thereby empowering more nuanced and fine-grained control during the text-guided generation.


\subsection{Cross-Modal Face Recognition}
\label{subsec:related_work_cmfr}
Face recognition is a longstanding and quintessential problem in the field of computer vision, which has witnessed substantial advancements over time. 
In its initial stages, traditional approaches rely on local descriptors (\eg, LBP~\cite{ahonenFaceDescriptionLocal2006}, HOG~\cite{denizFaceRecognitionUsing2011a}, SIFT~\cite{bicegoUseSIFTFeatures2006}) to extract face features. 
With the advent of deep convolutional neural networks (CNNs) and delicate design of loss functions~\cite{liuSphereFaceDeepHypersphere2017, wangCosFaceLargeMargin2018, dengArcFaceAdditiveAngular2019}, contemporary research has shifted towards utilizing these powerful frameworks to obtain superior performance and rapidly extended to concerns on cross-modal face recognition tasks.
By analyzing multi-modal facial features and mapping them to consistent latent space, these methods allow for the identification and categorization of individuals.
The Near-Infrared Spectrum (NIS) images and Visible Light Spectrum (VIS) images are regarded as two modalities, as demonstrated in~\cite{heLearningInvariantDeep2017, heWassersteinCNNLearning2018, saxenaHeterogeneousFaceRecognition2016}, whose discrepancies are managed through subspace learning employing deep neural networks and the Wasserstein distance.
By altering facial attributes, a 3D Morphable Model is used in~\cite{galeaForensicFacePhotosketch2017} to generate a large set of synthetic images that are then utilized to fine-tune a deep network, originally pre-trained on face photos, for face photo-sketch recognition through transfer learning.
\cite{mingCrossmodalPhotocaricatureFace2021} specifically focuses on the recognition of photo-caricature faces through the utilization of multi-task learning. 
Their approach incorporates a dynamic weights learning module that automatically assigns weights based on the significance of each task, which enables the network to allocate more attention to challenging tasks rather than simpler ones.
LDCTBP~\cite{koleyCrossmodalFaceRecognition2023} presents a simultaneous demonstration of the efficacy of handcrafted features in photo-sketch and NIS-VIS recognition by using discrete cosine transform as an effective local feature descriptor for illumination normalization.

Indeed, current cross-modal face recognition systems have not yet transcended the scope of visual data and comparable investigations concerning photo-text modalities are scarce.
This can be attributed to the inherent challenges involved in resolving the substantial disparity between linguistic and visual processing.
On the other hand, the availability of data sets containing accurate facial descriptions is significantly inadequate, making it challenging to effectively promote the corresponding research efforts.
The ongoing project, Face2Text~\cite{tantiFace2TextRevisitedImproved2022}, aims to assemble a dataset comprising natural language descriptions of human faces but its size remains relatively small, with its latest v2 version containing only \num[group-separator={,}]{10559} images and \num[group-separator={,}]{17022} corresponding descriptions. 
Several other databases designed for face synthesis, such as MM-CelebA-HQ~\cite{xiaTediGANTextGuidedDiverse2021} and CelebAText-HQ~\cite{sunMulticaptionTexttoFaceSynthesis2021a}, incorporate automatically generated or manually annotated natural descriptions based on CelebFaces Attributes (CelebA) dataset~\cite{liuDeepLearningFace2015}. 
Nevertheless, they are unsuitable for our intended investigations due to the inseparable mixture of identity-relevant (\eg, eyebrows, nose) and identity-irrelevant information (\eg, expression, accessories, makeup) in the descriptions.
That is to say, although multiple cross-modal algorithms have been developed, it is important to note that text-based face recognition is limited and warrants further attention.

\begin{figure}[t]
    \centering
    \includegraphics[width=\linewidth]{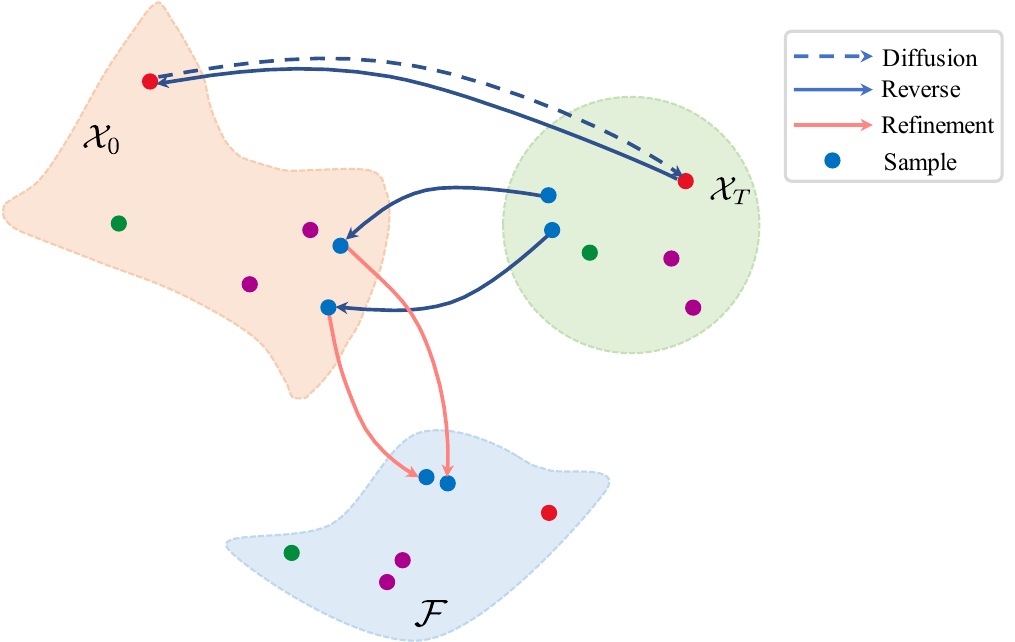}
    \caption{Theoretical depiction of probability density transport. Dots of the same color correspond to identical subjects. A random variable drawn from $\mathcal{X}_T$ is transported along the reverse path, contrary to the diffusion direction, to a sample subject to $\mathcal{X}_0$ through DPMs. Note that feature similarities are not explicitly regulated during this process. Ultimately, face recognition is accomplished through the refinement module, which further adjusts the feature distances within the space $\mathcal{F}$. See Sec.~\ref{subsec:Theoretical_Analysis} for details.}
    \label{fig:theory}
\end{figure}

\section{Method}
\label{sec:method}
The objective of our research endeavors is to expand the capabilities of diffusion models beyond generation, enabling them to accomplish text-to-image face recognition. 
In Sec.~\ref{subsec:Theoretical_Analysis}, we first provide a comprehensive analysis to establish the theoretical connection between DPMs and recognition problems, serving as the foundation of our framework.
The general problem formulation and specific algorithmic details pertaining to our methodology are subsequently presented in Sec~\ref{subsec:Problem_Formulation} and Sec.~\ref{subsec:algorithm},  respectively.

\begin{figure*}[t]
    \centering
    \includegraphics[width=\linewidth]{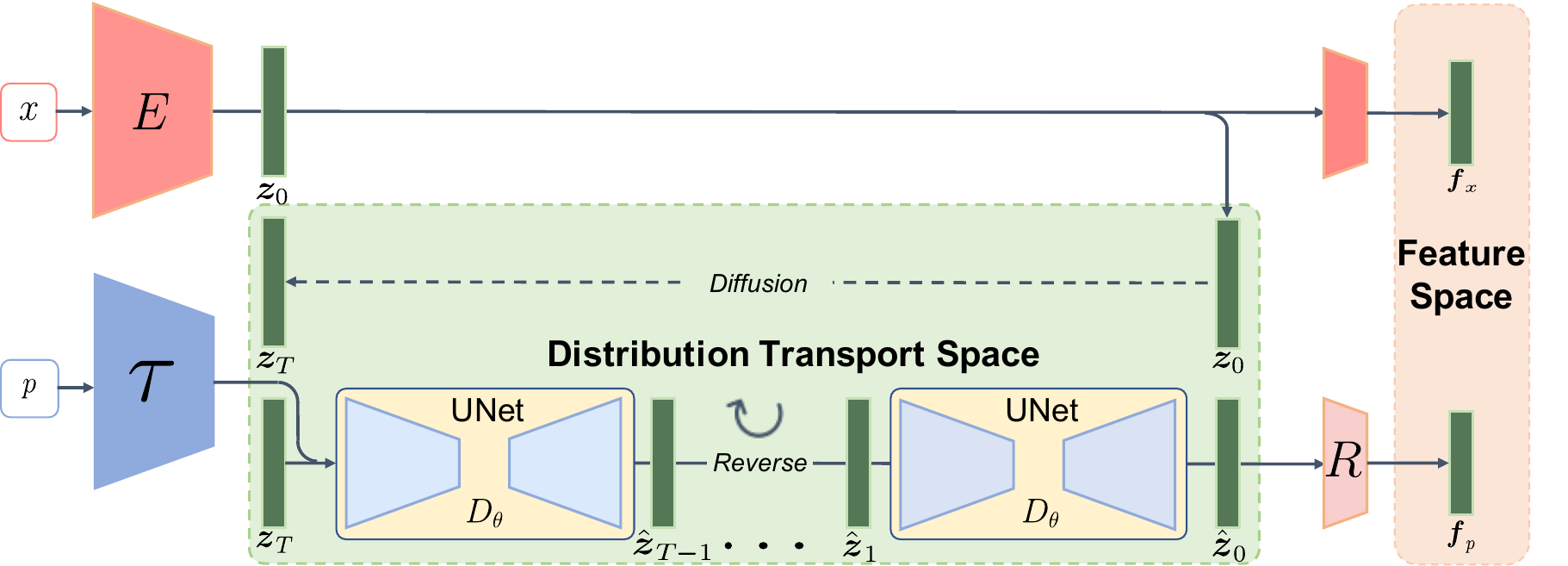}
    \caption{The complete training procedure and network architecture. As one of the bifurcated branches of encoder $E$, the intermediate feature $\boldsymbol{z}_0$ is extracted as the initial sample in the diffusion process. The diffusion model $D_\theta$, utilizing the UNet structure and taking the vectors tokenized by $\tau$ as inputs, is subsequently employed iteratively as the reverse of the diffusion process. Once $\hat{\boldsymbol{z}}_0$ is obtained, the refinement network $R$ maps it to $\boldsymbol{f}_p$ within the feature space $\mathcal{F}$. The final decision of the similarity between the facial image and the textual description is based on the distance between $\boldsymbol{f}_p$ and $\boldsymbol{f}_x$.}
    \label{fig:architecture}
\end{figure*}

\subsection{Theoretical Analysis}
\label{subsec:Theoretical_Analysis}
Let $\boldsymbol{z}_t$ represent a series of random variables indexed by time $t\in[0,T]$ in a diffusion process, then the initial samples $\boldsymbol{z}_0 \sim \mathcal{X}_0$, which exhibit independent and identically distributed (i.i.d) characteristics, will undergo an evolution leading to $\boldsymbol{z}_T \sim \mathcal{X}_T$ while gradually introducing additional noise.
The process is illustrated in Fig.~\ref{fig:theory} by the dashed arrow line, depicting the transition from samples in $\mathcal{X}_0$ to those in $\mathcal{X}_T$. 
Typically, $\mathcal{X}_T$ is a simple distribution to facilitate straightforward sampling during the reverse process.
It has been clarified in~\cite{songScoreBasedGenerativeModeling2021} that this forward diffusion process in DPMs can be formulated to the stochastic It\^o process:
\begin{equation}
\label{eq:ito}
    \mathrm{d} \boldsymbol{z}_t = \xi_t(\boldsymbol{z}_t) \mathrm{d}t + \sigma_t \mathrm{d}W_t,
\end{equation}
where $\xi_t$ and $\sigma_t$ are the drift and diffusion coefficient respectively, and $W_t$ is the standard Wiener process.
The Eq.~\eqref{eq:ito} signifies that the diffusion random variables under the Markov assumption is influenced by both deterministic and stochastic processes simultaneously.
This process has been demonstrated to be mathematically equivalent to the $n$-dimensional Fokker-Planck equation, which describes the partial derivative of the probability density $\rho_t$ with respect to time. 
The equation takes the following form:
\begin{equation}
\label{eq:fp}
    \frac{\partial \rho_t}{\partial t} = \nabla \cdot(\rho_t v_t) + D_t\Delta \rho_t,
\end{equation}
where $v_t$ is the time-varying velocity field and $D_t=\sigma^2_t/2$ denotes the diffusivity.
Eq.~\eqref{eq:fp} offers an alternative perspective for understanding the diffusion process. 
Rather than adopting the particle-centered viewpoint found in Eq.~\eqref{eq:ito}, it allows us to perceive diffusion as a transportation mechanism between probability distributions.
If the duration is sufficiently long, the distribution $\mathcal{X}_T$ will ultimately converge to a standard Gaussian distribution when $v_t$ is specifically selected for degradation in each step, irrespective of the initial distribution $\mathcal{X}_0$, which has been demonstrated in~\cite{hoDenoisingDiffusionProbabilistic2020}. 
The objective of generating data from random noise is accomplished by reversing this straightforward forward process, wherein the direction and magnitude of each step are predicted by DPMs.
In order to restore the initial distribution $\mathcal{X}_0$, diffusion models are trained to predict the disparity between $\mathcal{X}_{t-1}$ and $\mathcal{X}_t$, utilizing the provided values of $t$ and $\boldsymbol{z}_t$ as input.
To be specific, for the DDPM algorithm, given a group of random time step $t$ and sample $\boldsymbol{z}_0$, the diffusion model parameterized by $\theta$ (denoted by $D_\theta$) during training is essentially searching for
\begin{equation}
\label{eq:diffusion_model}
    \operatorname*{argmin}_\theta \E \left[ d\big(D_\theta(\boldsymbol{z}_t, t), d(\boldsymbol{z}_t, \boldsymbol{z}_{t-1})\big) \right],
\end{equation}
where $d(\cdot,\cdot)$ is the distance that is specifically detailed in Sec.~\ref{subsec:Problem_Formulation} and Sec.~\ref{subsec:algorithm}.
When $D_\theta$ is properly trained, the reverse of the diffusion process, initiated at $\hat{\boldsymbol{z}}_T$ which is sampled from $\mathcal{X}_T$ for generation, is accomplished through an iterative procedure described by
\begin{equation}
\label{eq:backward}
    \hat{\boldsymbol{z}}_{t-1} = f \big(\hat{\boldsymbol{z}}_t, D_\theta(\hat{\boldsymbol{z}}_t, t)\big).
\end{equation}
The Eq.~\eqref{eq:backward} demonstrates that $\hat{\boldsymbol{z}}_0$ can be obtained by some specific function $f$ when $\hat{\boldsymbol{z}}_T$ is given.
The reverse process is visually represented in Fig.~\ref{fig:theory} through blue arrow lines.
Under ideal conditions, it should be possible to to generate a sample subject to $\mathcal{X}_0$ by sampling from $\mathcal{X}_T$ distribution through multiple iterations, which is sufficient for unconditional generative algorithms.

While this approach successfully accomplishes the transportation from one distribution to another, thereby facilitating the resolution of cross-modal problems, it is not entirely appropriate for recognition tasks, whose benchmark is based on rigorous feature similarity.
In the task of face recognition, it is expected that the distances between inter-class embeddings should be noticeably greater than the distances between intra-class embeddings.
In fact, the explicit assurance of this requirement is not deemed necessary in the LDMs currently tailored for content generation.
Owing to their powerful decoders, LDMs possess the ability to produce satisfactory images, provided that the resulting sample $\hat{\boldsymbol{z}}_0$ approximately subjects to the distribution of $\mathcal{X}_0$.
However, it is imperative to emphasize that the distances of samples in $\mathcal{X}_0$ from the identical subject (\eg, blue dots) in Fig.~\ref{fig:theory} are not mandated to be closer than those of different subjects (\eg, the upper blue dot and purple dot).
The latent embeddings $\hat{\boldsymbol{z}}_0$ are consequently unsuitable for direct face recognition, a conclusion that is further substantiated through experimental evidence presented in Sec.~\ref{subsec:ablation}.

In order to ensure the viability of the framework for the face recognition task, we have undertaken the specific design of an additional network, denoted as $R$, with the purpose of further refining the rough estimate $\hat{\boldsymbol{z}}_0$ by mapping it into a more reasonable feature space referred to as $\mathcal{F}$.
When undergoing rearrangement through the application of $R$, the refined features $R(\hat{\boldsymbol{z}}_0)$ within $\mathcal{F}$, as depicted in Fig.~\ref{fig:theory}, are clustered based on their corresponding identities.
Details about the structure and implementation are clarified in Sec.~\ref{subsec:algorithm}.

\begin{algorithm}[t]
\caption{Training $D_\theta$}
\label{alg:D}
\INPUT Facial images $x$; description prompt $p$; encoder $E$; maximum time step $T$ \\
\OUTPUT Trained diffusion model $D_\theta$
    \begin{algorithmic}[1]
        \While {not converged}
        \State $\boldsymbol{z}_0 \gets E_z(x)$
        \State $t \sim U(\{1,...,T\})$
        \State $\epsilon \sim \mathcal{N}(0, \boldsymbol{I})$
        \State $\boldsymbol{z}_t \gets \sqrt{\bar{\alpha}_t} \boldsymbol{z}_0 + \sqrt{1-\bar{\alpha}_t} \epsilon$ 
        \State $L \gets \norm{D_\theta(\boldsymbol{z}_t,t,p)-\epsilon}^2$
        \State Update $\theta$ to reduce $L$
        \EndWhile
    \end{algorithmic}
\end{algorithm}

\subsection{Problem Formulation for Cross-Modal Face Recognition}
\label{subsec:Problem_Formulation}
After establishing the theoretical framework in Sec.~\ref{subsec:Theoretical_Analysis}, our subsequent focus is directed towards the specific problem of text-to-image face recognition. 
Based on the LDMs, it is reasonable to take the lower-dimensional i.i.d latent variables as $\boldsymbol{z}$, rather than higher-dimensional images in the original DPMs. 
Furthermore, $\mathcal{X}_0$ is considered to be reconstructed by $D_\theta$ from $\mathcal{X}_T$, guided by prompts denoted by $p$.
That is to say, the model $D_\theta$ is anticipated to predict the added noise through vectorized prompts $p$.
Since the diffusion process is deterministic, the series of $\boldsymbol{z}_t$ is able to be simply obtained when $\boldsymbol{z}_T$ sampled from Gaussian distribution $\mathcal{N}(0,\boldsymbol{I})$.
Taking $\boldsymbol{z}_t$, $t$ and $p$ as inputs, the loss function for training $D_\theta$ is
\begin{equation}
\label{eq:l_ldm}
    L_{LDM} = \E_{\boldsymbol{z}_0, t, p, \epsilon\sim\mathcal{N}(0,\boldsymbol{I})} \norm{D_\theta(\boldsymbol{z}_t, t, p) - \epsilon}^2,
\end{equation}
where $t$ is sampled from a uniform distribution $ U(\{1, ... ,T\})$ and $\norm{\cdot}$ is chosen to be the $\ell_2$ norm in this work.
In a manner akin to the process described in~\cite{rombachHighResolutionImageSynthesis2022a}, the initial variable $\boldsymbol{z}_0$ undergoes degradation to yield $\boldsymbol{z}_t$ through a function that is associated with the noise $\epsilon$, employing the reparameterization trick.

Specifically, we employ a pretrained CNN-based network, denoted as $E$, which possesses sufficient capabilities in conventional face recognition, as the encoder.
Given an input image $x$ in RGB space and the encoder $E$, the corresponding feature denoted by $\boldsymbol{f}_x$ for recognition is 
\begin{equation}
    \boldsymbol{f}_x = E(x) = E_f\big(E_z(x)\big),
\end{equation}
where the encoder $E$ is divided into two branches, namely $E_z$ and $E_f$.
In our work, $E_z(x)$ serves as the initial sample $\boldsymbol{z}_0$ for recovery during the training of $D_\theta$, which means
\begin{equation*}
    \boldsymbol{z}_0 = E_z(x).
\end{equation*}
This bifurcation is also clearly described in Fig.~\ref{fig:architecture}.

To achieve the appropriate mapping from $\mathcal{X}_0$ to $\mathcal{F}$, the refinement network $R$ is trained after the completion of training of $D_\theta$. 
We utilize the cosine embedding loss to train $R$, meaning that the loss function $L_R$ is defined as
\begin{equation}
\label{eq:l_r}
    L_R = \E_{\hat{\boldsymbol{z}}_0} \left[ \frac{R(\hat{\boldsymbol{z}}_0)}{\norm{R(\hat{\boldsymbol{z}}_0)}} \cdot \frac{E(x)}{\norm{E(x)}} \right],
\end{equation}
where $\hat{\boldsymbol{z}}_0$ is attained through the iteration described in Eq.~\eqref{eq:backward}, given a fix time step $t$ and textual description $p$.
Finally, the textual feature based on the description of a facial image is obtained by
\begin{equation*}
    \boldsymbol{f}_p = R(\hat{\boldsymbol{z}}_0).
\end{equation*}
The rest layers $E_f$ of $E$ progressively encode $\boldsymbol{z}_0$ into $\boldsymbol{f}_x$, enabling a conclusive comparison with the text-based feature $\boldsymbol{f}_p$ within the feature space $\mathcal{F}$ for the purpose of recognition.

\begin{algorithm}[t]
\caption{Training $R$}
\label{alg:R}
\INPUT Facial images $x$; description prompt $p$; inference steps $\Tilde{T}$; encoder $E$;\\
\OUTPUT Trained refinement network $R$
    \begin{algorithmic}[1]
        \While {not converged}
        \State $\boldsymbol{z}_T \sim \mathcal{N}(0, \boldsymbol{I})$
        \For {$t=\Tilde{T},...,1$}
            \State $\hat{\boldsymbol{z}}_{t-1} \gets$ Eq.~\eqref{eq:sampling}
        \EndFor
        \State $L \gets \frac{R(\hat{\boldsymbol{z}}_0)}{\norm{R(\hat{\boldsymbol{z}}_0)}} \cdot \frac{E(x)}{\norm{E(x)}}$
        \State Update parameters of $R$ to reduce $L$
        \EndWhile
    \end{algorithmic}
\end{algorithm}

\subsection{Algorithm and Architecture}
\label{subsec:algorithm}
In this section, we present comprehensive information regarding the complete design of algorithms and structures illustrated in Fig.~\ref{fig:architecture}.
The encoder $E$ used in this study is a conventional face recognition network constructed by ResNet~\cite{heDeepResidualLearning2015}, incorporating the marginal loss proposed by ArcFace~\cite{dengArcFaceAdditiveAngular2019}. 
Initially, we train the model on the specific task of face recognition with pure facial images $x$, until both $\boldsymbol{f}_x$ and $\boldsymbol{z}_0$ achieve a significantly high level of accuracy.
The parameters of the encoder are then completely fixed throughout all subsequent procedures. 

Once the encoder is adequately prepared, $\boldsymbol{z}_0$, the output of the intermediate layers, is obtained by $E_z(x)$ and utilized for crucial training on $D_\theta$ in the continuous steps.
The tokenizer, denoted as $\tau$, serves as the initial step in the diffusion process, transforming the prompts $p$ into vectors.
The main outline of the algorithm for training $D_\theta$ is shown in Alg.~\ref{alg:D}.
Given the step $t$ sampled from a uniform distribution $U(\{1,...,T\})$ and the noise $\epsilon$ sampled from standard normal distribution $\mathcal{N}(0, \boldsymbol{I})$, the diffused product $\boldsymbol{z}_t$ at step $t$ is given by
\begin{equation}
\label{eq:ddpm}
    \boldsymbol{z}_t = \sqrt{\bar{\alpha}_t} \boldsymbol{z}_0 + \sqrt{1-\bar{\alpha}_t} \epsilon,
\end{equation}
where $\bar{\alpha}_t$ is a hyperparameter controlling the added noise in each step of diffusion process.
In fact, Eq.~\eqref{eq:ddpm} is equivalent to 
\begin{equation*}
    \boldsymbol{z}_t = \sqrt{\alpha_t} \boldsymbol{z}_{t-1} + \sqrt{1-\alpha_t}\epsilon,
\end{equation*}
which indicates $\boldsymbol{z}_t$ is sampled from the normal distribution $\mathcal{N}(\sqrt{\alpha_t}\boldsymbol{z}_{t-1}, (1-\alpha_t) \boldsymbol{I})$ with $\bar{\alpha}_t=\prod_i^t \alpha_i$.
The loss function described in Eq.~\eqref{eq:l_ldm} is then computed using the prompts $p$ and the corresponding $\boldsymbol{z}_t$, which are assigned to the variable $L$ in Alg.~\ref{alg:D} to realize the optimization process.

\begin{figure*}[t]
    \centering
    \subfloat[]{\includegraphics[width = 0.48\linewidth]{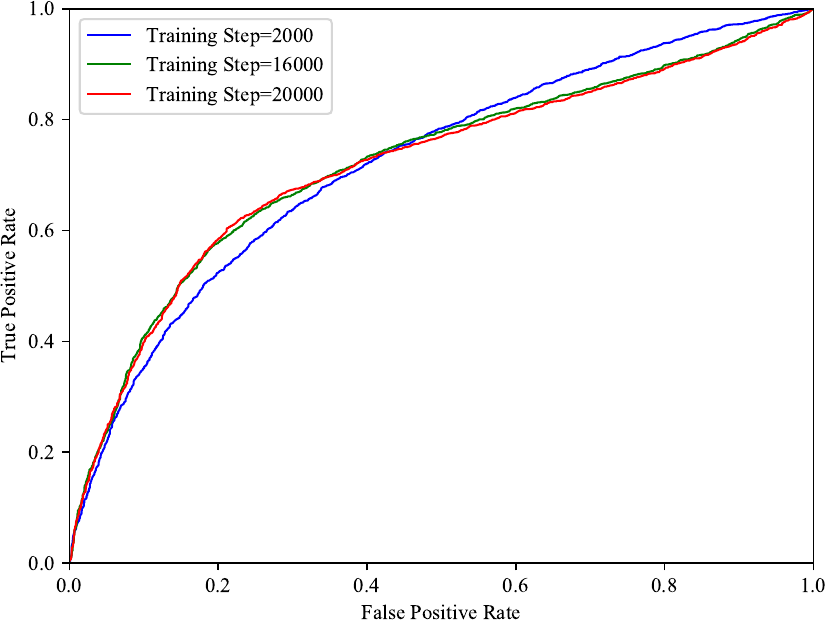}\label{fig:valid_acc_step}}
    \hfill
    \subfloat[]{\includegraphics[width = 0.48\linewidth]{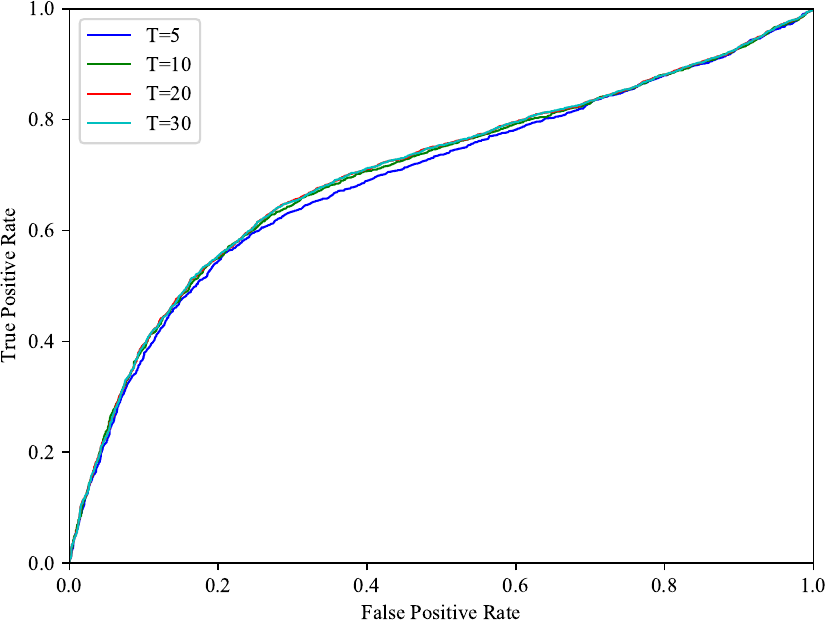}\label{fig:valid_time}}
    \caption{Intermediate observations during training steps evaluated on the validation set.}
    \label{fig:validation}
\end{figure*}

Before training the refinement network $R$, we need to iteratively sample $\hat{\boldsymbol{z}}_0$, which is initialized from $\boldsymbol{z}_T$, after fixing the parameters of $D_\theta$.
During the sampling process, the parameter $\Tilde{T}$ is designated as a hyperparameter that determines the length of inference step.
We designate the variable $p$ as one of the direct inputs to $D_\theta$, omitting the presence of the tokenizer $\tau$ for the sake of simplicity.
Similar to the approach employed in DDPM~\cite{hoDenoisingDiffusionProbabilistic2020}, the sampling function $f$ in Eq.~\eqref{eq:backward} is executed through the following procedure:
\begin{equation}
\label{eq:sampling}
    \boldsymbol{z}_{t-1} = \frac{1}{\sqrt{\alpha_t}} \left( \boldsymbol{z}_t-\frac{\beta_t}{\sqrt{1-\bar{\alpha}_t}} D_\theta(\boldsymbol{z}_t, t, p) \right) + \sigma_t \eta,
\end{equation}
where $\beta_t=1-\alpha_t=\sigma_t^2$ with $\eta\sim \mathcal{N}(0, \boldsymbol{I})$ for $t>1$ and $\eta=0$ for $t=1$.
We proceed to train the refinement network $R$ using the cosine similarity defined in Eq.~\eqref{eq:l_r}, based on the values of $\boldsymbol{z}_0$ that have already been obtained through this sampling procedure.
The features obtained from a sufficiently trained encoder $E$ exhibit reduced inner-class distances compared to inter-class distances, making them particularly suitable for face recognition. 
Due to the incorporation of $E(x)$ as a guiding factor, the refinement module $R$ successfully transforms the space $\mathcal{X}_0$ into $\mathcal{F}$ by employing a lightweight network consisting solely of PReLU~\cite{heDelvingDeepRectifiers2015} and linear layers. 
This effectiveness is further supported by the experimental outcomes presented in Sec.~\ref{sec:experiments}.

\begin{table}[t]
    \caption{Detailed partitions on CelebA. \\ The quantity is displayed in each cell.}
    \centering
    \begin{tabular}{l c c}
    \toprule
     Purpose    & Subjects & Images \\
     \midrule
     Training   &  $5000$  & $96097$ \\
     Validation &  $2000$  & $40242$ \\
     Test       &  $3177$  & $61260$ \\
    \bottomrule
    \end{tabular}
    \label{tab:dataset}
\end{table}

\section{Experiments}
\label{sec:experiments}
In this section, we present a thorough panorama of the experiments conducted on our cross-modal DiFace model, designed to achieve text-to-face recognition, through rigorous evaluations. 
The experimental settings in Sec.~\ref{subsec:setup}, encompassing the utilized datasets, the benchmark criteria and the specific parameters, are serving as the foundation for subsequent analysis.
Results of evaluations and analyses of our algorithm are impartially presented in Sec.~\ref{subsec:evaluation} and Sec.~\ref{subsec:ablation}.
In Sec.~\ref{subsec:visualization}, we additionally provide visualized examples to elucidate the inherent difficulties arising from the intrinsic imprecision of verbal descriptions compared to texture information. 
These challenges cannot be surmounted by algorithms.

\subsection{Experimental Setting}
\label{subsec:setup}
\subsubsection{Datasets} 
Existing facial image databases containing a vast number of images paired with corresponding identity labels and descriptive metadata are considered insufficient, as discussed in further detail in Sec.~\ref{subsec:related_work_cmfr}. 
Given the circumstances, the CelebA dataset employed for the evaluation could be considered the most suitable option for this research owing to its extensive compilation of \num[group-separator={,}]{202599} celebrity images, each annotated with \num[group-separator={,}]{40} binary attributes. 
We subsequently transform a portion of annotations related to identity features into linguistic prompts, and partition the dataset into distinct training, validation and test sets without any intersection. 
The scale of each component of the reorganized CelebA is summarized in Tab.~\ref{tab:dataset}.
Furthermore, a subset of the WebFace~\cite{zhuWebface260mBenchmarkUnveiling2021} is applied for the pretrained face recognition encoder $E$.
AgeDB~\cite{moschoglouAgeDBFirstManually2017}, LFW~\cite{huangLabeledFacesWild2008}, CPLFW~\cite{zhengCrossPoseLFWDatabase}, CALFW~\cite{zhengCrossAgeLFWDatabase2017}, CFP-FF/FP~\cite{senguptaFrontalProfileFace2016} are used for ablation study.

\subsubsection{Criteria}
We employ a verification (1:1) approach to assess the performance of DiFace, as indicated by the success rate of accurately predicting the positive or negative pairs within the test set.
To be more specific, each of the cosine similarity $\mathcal{S}(x, p)$ between normalized feature of images and prompts is obtained through 
\begin{equation}
    \mathcal{S}(x, p) = \frac{\boldsymbol{f}_x \cdot \boldsymbol{f}_p}{\norm{\boldsymbol{f}_x} \norm{\boldsymbol{f}_p}}.
\end{equation}
The $i$-th Boolean prediction $\Gamma(x, p)_i$ is defined
\begin{equation}
\label{eq:threshold}
    \Gamma(x, p)_i = \left\{ 
                \begin{array}{ll}
                   1 & \mathcal{S}(x, p) \geq s, \\
                   0 & \mathcal{S}(x, p) < s
                \end{array}
    \right. ,
\end{equation}
where $s$ is the threshold defined according to the best performance in the validation set.
More discussions about the details of $s$ are provided in Sec.~\ref{subsubsec:training_procedure}.
Let $i$ represent the index of the $i$-th pair $(x, p)$ sampled from the test set, where a total of $N$ pairs are considered. 
If $y_i$ denotes the ground truth label, wherein it takes the value of $1$ only if both $x$ and $p$ are chosen from the same identity, and $0$ otherwise, the accuracy rate $r$ can be straightforwardly given by
\begin{equation}
\label{eq:accuracy}
    r = \frac{1}{N} \sum_{i=1}^N \mathbbm{1}(\Gamma(x, p)_i = y_i),
\end{equation}
where $\mathbbm{1}$ represents the indicator function.

\begin{figure*}[t]
    \centering
    \subfloat[]{\includegraphics[width = 0.48\linewidth]{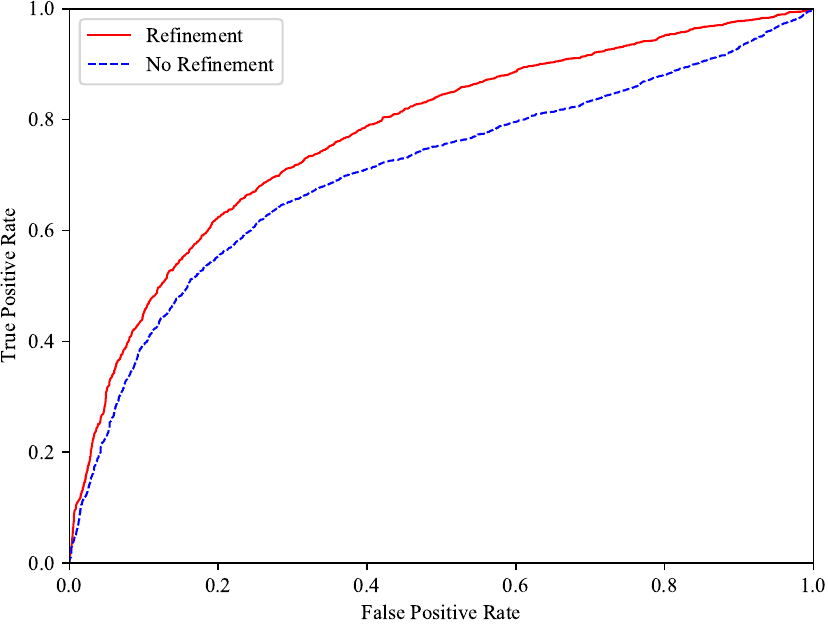}\label{fig:valid_roc}}
    \hfill
    \subfloat[]{\includegraphics[width = 0.48\linewidth]{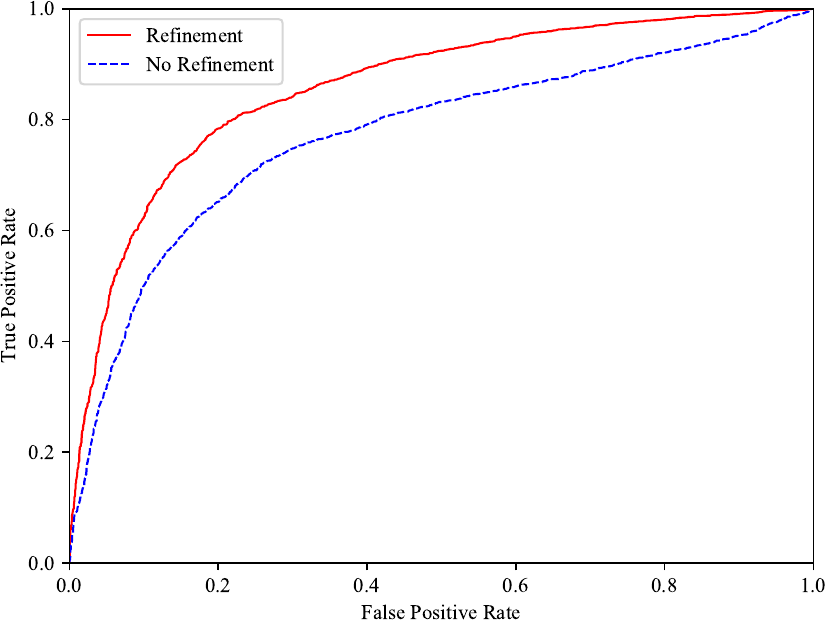}\label{fig:test_roc}}
    \caption{ROC curves of the optimal model on the validation (a) and test (b) set.}
    \label{fig:roc}
\end{figure*}

We additionally perform the more challenging identification (1:N) benchmark to further investigate the efficacy of DiFace.
The gallery set consists of images depicting various different subjects, while the corresponding prompts serve as the probe. 
Specifically, the feature similarity between each description $p_i$ indexed by $i$ in the probe set and all $N$ images $x_j$ indexed by $j$ in the gallery set is calculated, and the top $k$ scores are ranked.
The indexes of the top $k$ scores constitute the collection $c_i^k$, which is defined by
\begin{equation}
    c_i^k = \operatorname*{argmax}_{I \subset [N]: |I| = k } \sum_{j \in I} \mathcal{S}(x_j, p_i), 
\end{equation}
where $[N] = \{1,2,\ldots, N\}$.
In our work, it is chosen as the ground truth that the $i$-th paired images and prompts, denoted as $(x, p)_i$, correspondingly describe the identical subject.
The prediction, represented by $\gamma(x, p)_i$, is defined by
\begin{equation}
\label{eq:identification_accuracy}
    \gamma(x, p)_i = \left\{ 
                \begin{array}{ll}
                   1 & i \in c_i^k, \\
                   0 & i \notin c_i^k
                \end{array}
    \right. .
\end{equation}
Similar to Eq.~\eqref{eq:accuracy}, the accuracy for identification task is 
\begin{equation}
    r = \frac{1}{N} \sum_{i=1}^N \mathbbm{1}(\gamma(x, p)_i = 1).
\end{equation}
Due to the interference of erroneous images within the gallery, the task of identification becomes considerably more challenging when contrasted with verification. 
The value of both benchmarks, accompanied by relevant experiments, is extensively deliberated in Sec.~\ref{subsec:evaluation}.

\subsubsection{Parametric settings}
The facial images in RGB channels utilized in this study undergo alignment, cropping and resizing to achieve a resolution of $112 \times 112$ pixels. 
The final evaluated DiFace model is trained starting from a learning rate of $1\times10^{-4}$ and a mixed precision of BF$16$. 
Both the batch size and gradient accumulation steps are set to $4$ with the maximum gradient norm restricted to $1$.
We incorporate the exponential moving average (EMA) technique for the models' weights, along with an $8$-bit Adam optimizer.
Additionally, the feature scaling factor and the feature embedding dimension are applied to $0.3$ and $512$ respectively.

\subsection{Evaluations on DiFace}
\label{subsec:evaluation}
In this section, we present a comprehensive account of the experimental procedure and provide a thorough analysis of the results obtained.

\begin{figure}[t]
    \centering
    \includegraphics[width=\linewidth]{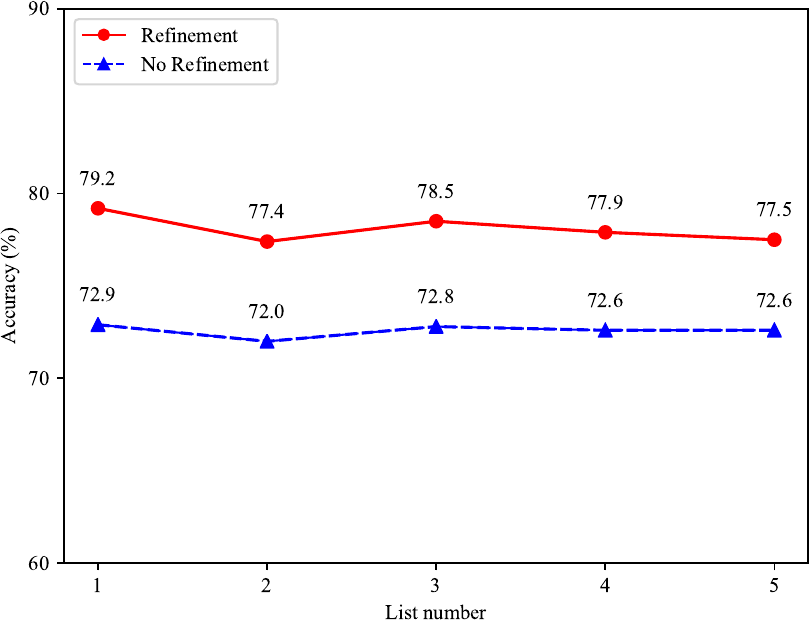}
    \caption{The verification accuracy is evaluated on five different pair lists randomly extracted from the test set. The consistently high and stable accuracy demonstrated by our model indicates its reliable performance, suggesting that the test data is unbiased without any filtration.}
    \label{fig:verification}
\end{figure}

\subsubsection{Training procedure}
\label{subsubsec:training_procedure}
Driven by algorithms in Sec.~\ref{subsec:algorithm}, the controlled diffusion model is initially trained to achieve convergence, as evidenced by the reduction in loss and the resultant improvement in validation accuracy. 
The determination of thresholds and their corresponding accuracies is founded upon the receiver operating characteristic (ROC) curve depicted in Fig.~\ref{fig:validation}. 
In this graphical representation, the axes represent the true positive rate (TPR) and false positive rate (FPR).
The values of TPR and FPR fluctuate relative to the threshold $s$, making them functions of $s$ that can be denoted as $T(s)$ and $F(s)$.
As the number of training steps increases, the ROC curve in Fig.~\ref{fig:valid_acc_step} demonstrates improvement.
The determination of the threshold for the similarity score $s$ in Eq.~\eqref{eq:threshold} is achieved in detail by maximizing the expression:
\begin{equation*}
    s = \operatorname*{argmax}_s T(s)-F(s).
\end{equation*}
Based on the line chart depicted in Fig.~\ref{fig:valid_acc_step}, we opt to select the checkpoint at step \num[group-separator={,}]{20000} to undergo a finetune will learning rate of $5\times10^{-5}$ before the final decision to ensure the stability and reliability of subsequent experiments.

The designated step $\Tilde{T}$ for inference significantly impacts the performance of recognition, as illustrated in Fig.~\ref{fig:valid_time}. 
Excessively large values of $\Tilde{T}$ result in increased inference time, while excessively small values of $\Tilde{T}$ lead to a decrease in accuracy.
Therefore, we tend to choose a predetermined time series that strikes a balance between efficiency and effectiveness.
Based on these considerations and the experimental results depicted in Fig.~\ref{fig:valid_time}, we ultimately determine that the value of $\Tilde{T}$ should be set to $20$ during the subsequent tests..

Once the training process of the diffusion model is completed, we proceed to undertake the individual training on the refinement model according to Alg.~\ref{alg:R}.
During the process of enhancing the capacity of the refinement model, the distance between the feature embeddings $R(\hat{\boldsymbol{z}}_0)$ and $E(x)$ is continuous reduced.
The resultant ROC curves for face verification, represented by the red line, are prominently illustrated in Fig.~\ref{fig:valid_roc} to showcase the performance of the optimal model in both the validation and test datasets. 
Additionally, Fig.~\ref{fig:valid_roc} provides evidence that the refinement module effectively enhances recognition performance, a topic further elaborated by the ablation study in Sec.~\ref{subsec:ablation}.

\begin{figure}[t]
    \centering
    \includegraphics[width=\linewidth]{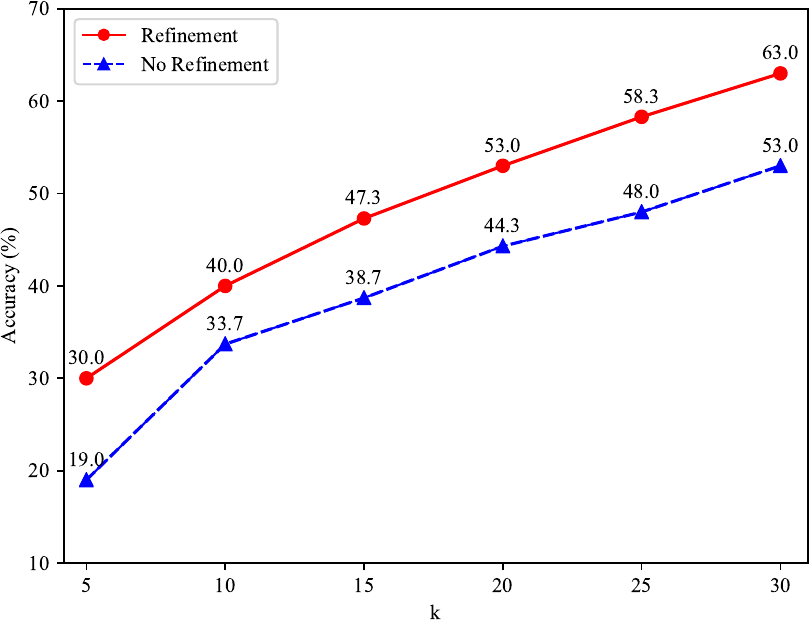}
    \caption{The identification accuracy with respect to $k$ is evaluated on the pair list randomly extracted from the test set. The accuracy curves manifest the substantial efficacy of our model in discerning and eliminating irrelevant images solely based on textual descriptions.}
    \label{fig:identification}
\end{figure}

\subsubsection{Results of face verification}
During the testing phase, we employ a random selection process to compile a list comprising \num[group-separator={,}]{12000} paired facial images and corresponding description prompts. 
It is noteworthy that half of these pairs belong to the same identity, while the remaining pairs involve distinct identities.
In order to mitigate the occurrence of chance factors, these procedures are repeatedly executed to yield a total of five distinct lists in the final assessments.
The success rates of verification for all pair lists in the benchmark are depicted in Fig.~\ref{fig:verification} in order to present the stability of our model.
Fig.~\ref{fig:verification} further demonstrates that our DiFace approach has attained a remarkable level of accuracy of nearly $80\%$ in text-to-face recognition, surpassing mere stochastic effectiveness.
Besides, the red line in Fig.~\ref{fig:test_roc} also displays the ROC result for the paired list of number 1.

Indeed, accurately identifying a face solely based on a few linguistic prompts is a challenging task, even for humans, within the realm of reality.
Given the substantial disparity between textual descriptions and visual images, our algorithm has demonstrated impressive performance, exhibiting a sufficiently high level of accuracy.
To be precise, it is possible for two distinct individuals to possess comparable descriptions or even identical prompts, owing to the limited annotations present within the database.
Considering this constraint, it is inevitable that a portion of misidentifications will occur, thus making it reasonable for the theoretical upper limit of text-to-image face recognition to be significantly below $100$ percent.
The performance of DiFace has demonstrated its capacity for verification in situations where image-image matches are prohibited.

\subsubsection{Results of face identification}
In accordance with the criteria outlined in Sec.~\ref{subsec:setup}, we conduct the more challenging task of face identification to evaluate the DiFace model. 
A total of $300$ images and their corresponding textual descriptions are randomly chosen from the test set to form a list.
Each pair originates from a distinct subject, indicating that the remaining $299$ pairs are considered noise in relation to each individual pair.
For every defined $k$, the accuracy is computed in accordance with Eq.~\eqref{eq:identification_accuracy}, and the outcomes are presented in Fig.~\ref{fig:identification}.
The accuracy rate $r$ represented by the red line increases as $k$ increases, as a larger value of $k$ indicates a wider range for the matching between texts and facial images.

\begin{figure*}[t]
    \centering
    \subfloat[AgeDB]{\includegraphics[width = 0.31\linewidth]{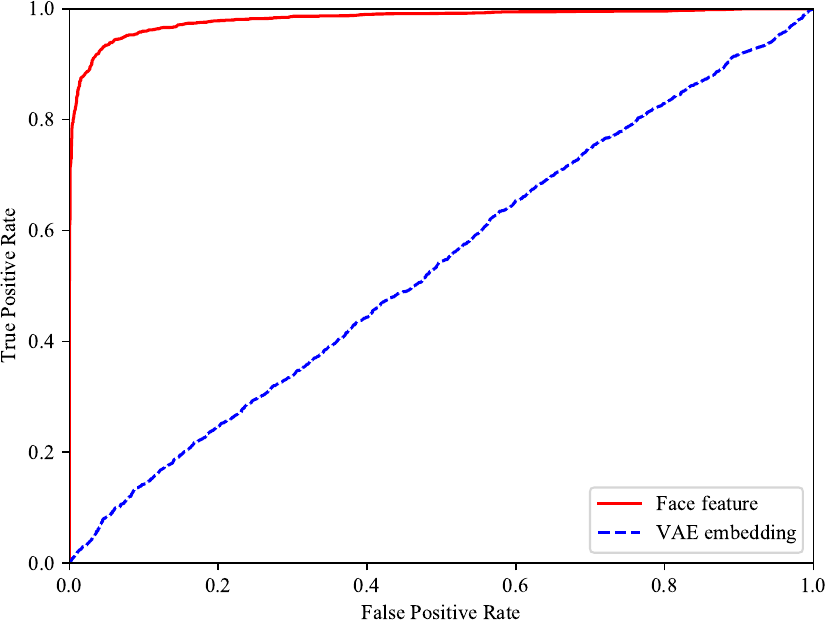}}
    \hfill
    \subfloat[LFW]{\includegraphics[width = 0.31\linewidth]{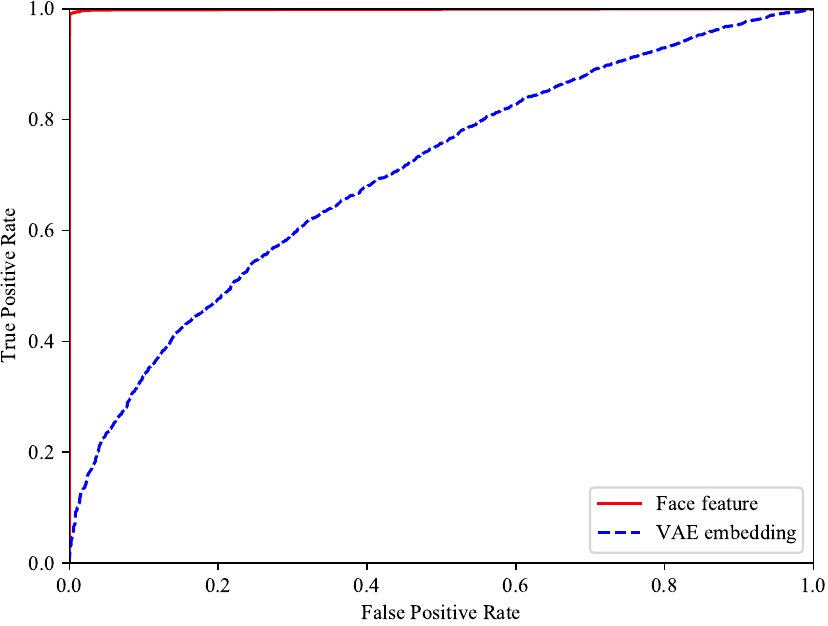}}
    \hfill
    \subfloat[CPLFW]{\includegraphics[width = 0.31\linewidth]{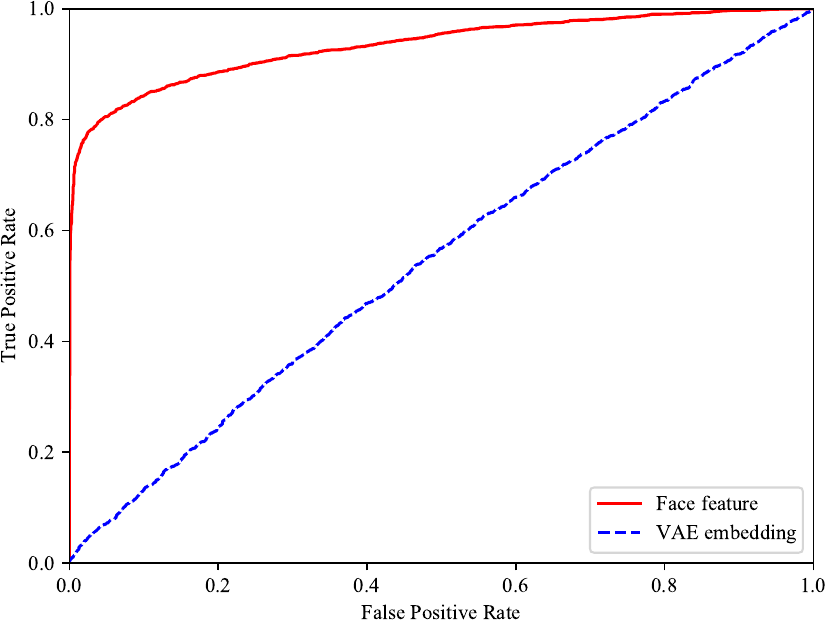}} \\
    \subfloat[CALFW]{\includegraphics[width = 0.31\linewidth]{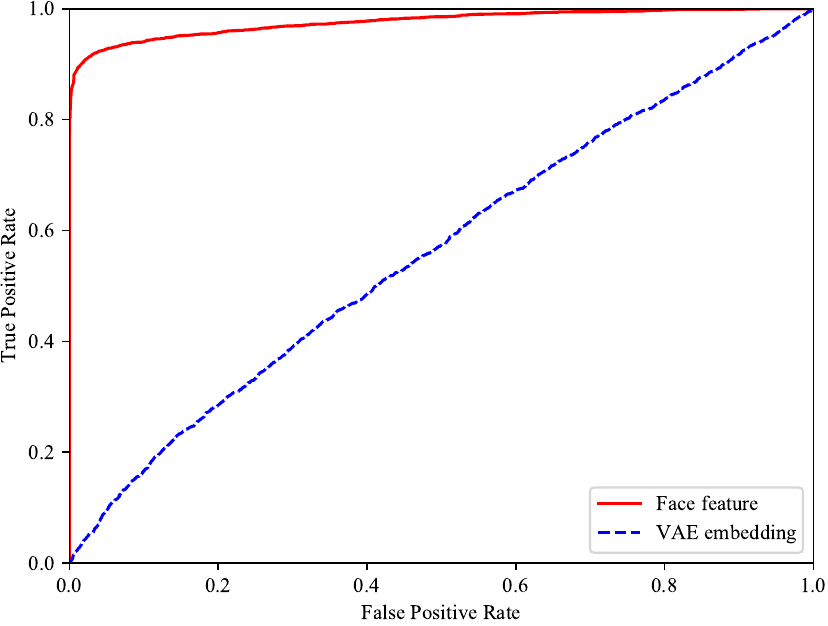}}
    \hfill
    \subfloat[CFP-FF]{\includegraphics[width = 0.31\linewidth]{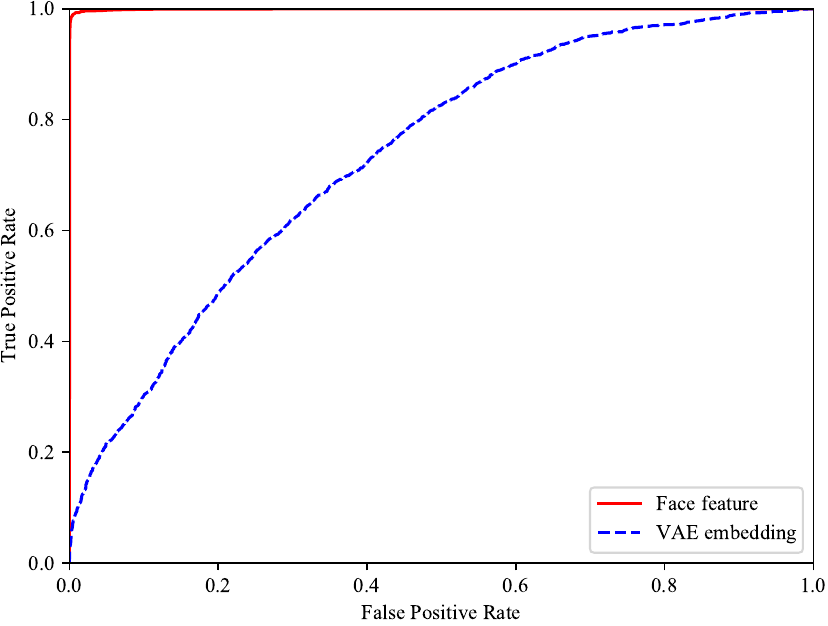}}
    \hfill
    \subfloat[CFP-FP]{\includegraphics[width = 0.31\linewidth]{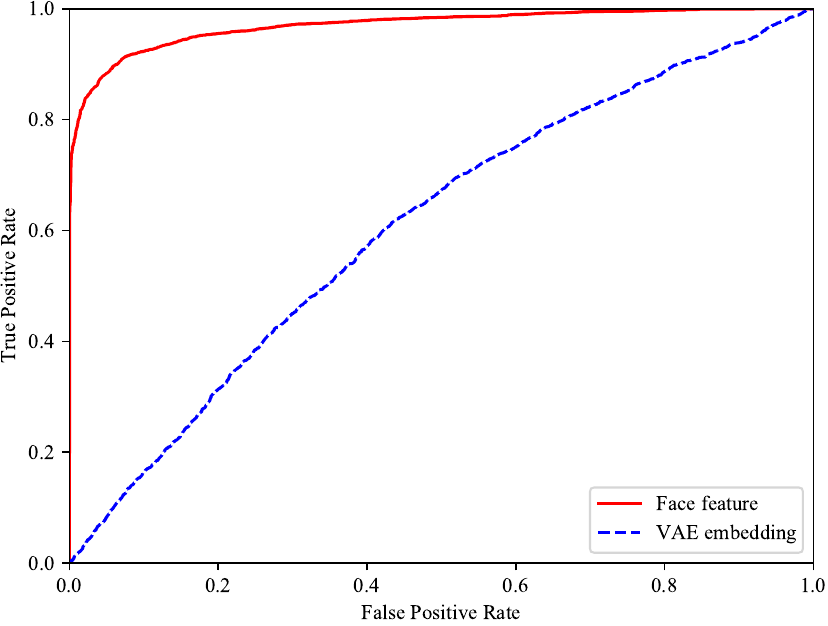}}
    \caption{The ROC curves representing the distinctive verification abilities of the face recognition model and VAE across 6 databases including AgeDB, LFW, CPLFW, CALFW, CFP-FF, CFP-FP.}
    \label{fig:ablation_encoder}
\end{figure*}

Based on Fig.~\ref{fig:identification}, we can draw the conclusion that our framework holds significant value in terms of filtering, particularly in scenarios where image-image matching is prohibited. 
To elaborate further, DiFace effectively eliminates a substantial number of incorrect candidates with a commendable level of accuracy. 
For instance, in the experiment where $k$ is set to $30$, the model provides a precise prediction accuracy of $63\%$ when excluding $90\%$ of interference candidates.

In circumstances where only textual descriptions are provided, our algorithm exhibits exceptional proficiency in discerning and eliminating inconsequential facial images, attaining a remarkably high level of accuracy. 
This achievement is particularly noteworthy when considering the inherent limitations of textual prompts, which lack the informative richness and contextual nuances present in visually-driven images.

\subsection{Ablation Study}
\label{subsec:ablation}
In order to conduct a comprehensive analysis of our DiFace model, we undertake an ablation study that specifically focuses on two modules, namely the refinement network $R$ and the encoder $E$.

\subsubsection{Refinement network}
We have presented a theoretical analysis regarding the rationale behind the implementation of the refinement network as discussed in Sec.~\ref{subsec:Theoretical_Analysis}. 
By effectively mapping samples from the space $\mathcal{X}_0$ to features in $\mathcal{F}$, we achieve enhanced clustering of embeddings, thereby improving the performance of face recognition.
In this section, we perform experiments to validate the aforementioned theoretical analysis through empirical evidence.

We exhibit the recognition performance using the similarity scores of $\boldsymbol{z}_0$ and $\hat{\boldsymbol{z}}_0$, independent of the involvement of the refinement module $R$. 
The depiction of this performance is represented by the blue dashed lines in Fig.~\ref{fig:roc}, Fig.~\ref{fig:verification} and Fig.~\ref{fig:identification}. 
In contrast, the recognition performance based on the scores of $\boldsymbol{f}_x$ and $\boldsymbol{f}_p$ is illustrated by the red lines for comparison.
The observation from both figures reveals that the complete model, encompassing the refinement network $R$, significantly outperforms the conditions in which $R$ is absent.

To be more precise, the ROC curve of feature $\boldsymbol{f}$ exhibits greater elevation compared to that of the latent variable $\boldsymbol{z}$ in Fig.~\ref{fig:roc}. 
This observation implies that employing $R$ for verification purposes results in a higher TPR compared to $\boldsymbol{z}$ at an equivalent FPR.
Moreover, the average accuracy in Fig.~\ref{fig:verification} is enhanced by an additional $5.52$ percentage points when employing the refinement network $R$ for the face verification task.
Likewise, the findings depicted in Fig.~\ref{fig:identification} further affirm that the refinement module, chosen based on its performance in the verification benchmark within the validation set, exhibits commendable efficacy in the context of the identification task as well.

In conclusion, the significance of the refinement network, as discussed in Sec.~\ref{subsec:Theoretical_Analysis}, has been substantiated through those experiments in Sec.~\ref{subsec:evaluation}.
Without the inclusion of this module, the pure diffusion network solely accomplishes the transformation of probability density from the language space to the latent space $\mathcal{X}_0$. 
The actual recognition process is ultimately achieved through the utilization of $R$, which adjusts the feature distance specifically tailored for face recognition.

\begin{figure*}
    \centering
    \includegraphics[width=\linewidth]{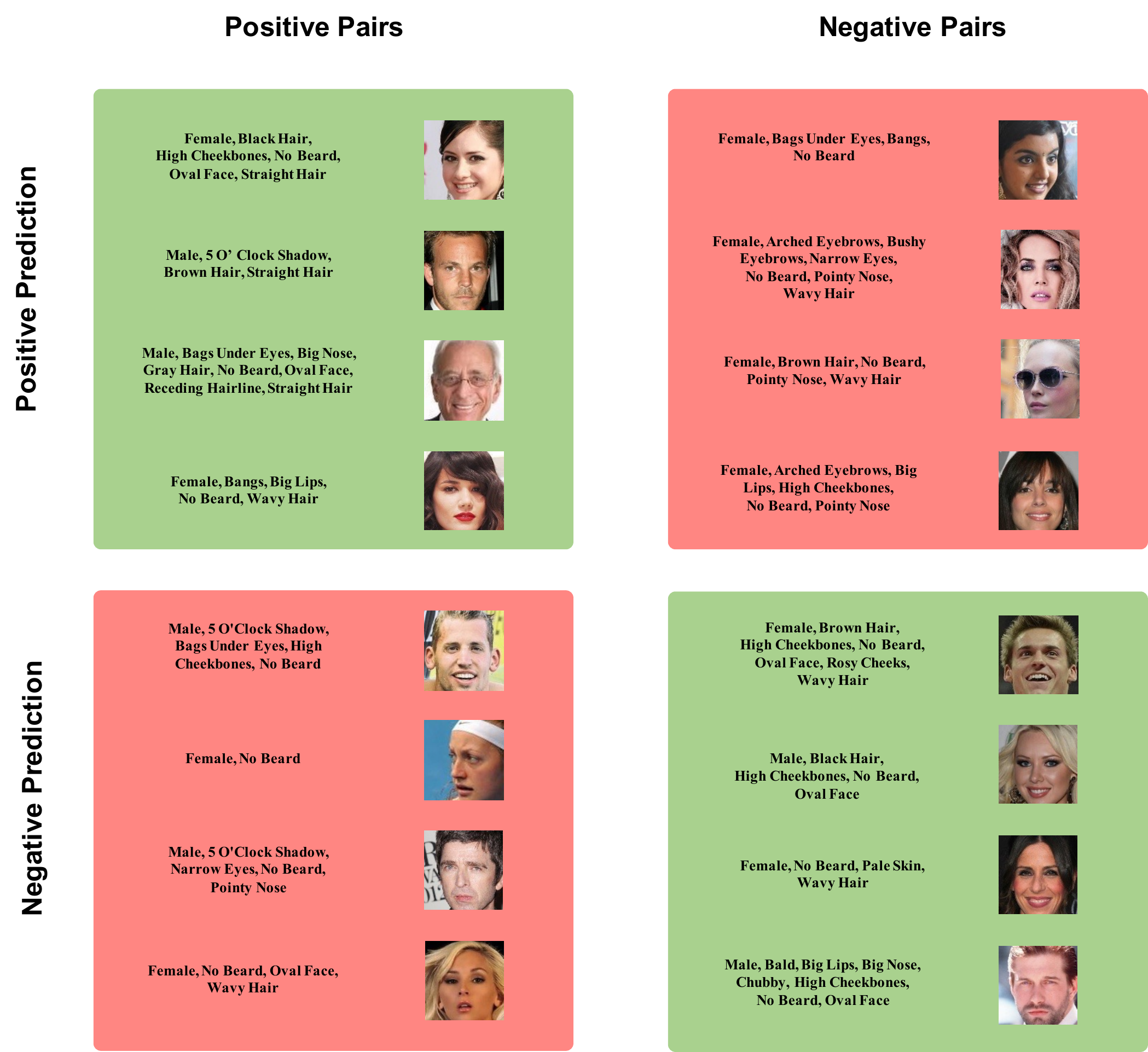}
    \caption{Visualization of face verification. Correct predictions are distinguished by a green background, while incorrect ones are marked with a red background.}
    \label{fig:visualization_verification}
\end{figure*}

\subsubsection{Encoder}
In the original framework of SD, a VAE structure assumes the role of an encoder responsible for encoding images into embeddings within a latent space. 
These embeddings can be regarded as features in some sense, as they maintain a vital connection to the input images, enabling the subsequent reconstruction of the images through the decoder.
However, it is important to note that despite their effectiveness in generative tasks, such embeddings are inherently unsuitable for recognition purposes. 
This limitation has been discussed in Sec.~\ref{subsec:Theoretical_Analysis}, where a comprehensive theoretical analysis has been presented.
Therefore, in this subsection, we present experimental evidence that highlights both the necessity and feasibility of utilizing a pretrained face recognition network as the dedicated encoder for such purposes.

The experiment is conducted by directly comparing the recognition ability between VAE-encoded embeddings and feature vectors derived from the face recognition model employed in this study.
We employ the identical VAE encoder that is utilized in the work of SD, in addition to our face recognition model which is trained using the technique introduced by~\cite{dengArcFaceAdditiveAngular2019}.
Based on the ROC curves depicted in Fig.~\ref{fig:ablation_encoder}, it is evident that the face features represented by the red lines exhibit superior performance compared to the VAE-encoded embeddings represented by the blue dashed lines.

It is important to note that the VAE-encoded embeddings play a crucial role as the target for the original diffusion models in SD to reverse. 
Due to their inherent limitations in face recognition, if the original VAE encoder is retained for training our framework, the whole performance will be further compromised.
Moreover, the findings additionally suggest that the embeddings within the sapce $\mathcal{X}_0$ are unsuitable for the purpose of face recognition when compared to the clustered feature vector in $\mathcal{F}$. This observation again strengthens the evidence presented in Sec.~\ref{subsec:Theoretical_Analysis}.

\subsection{Visualization}
\label{subsec:visualization}

\begin{figure*}
    \centering
    \includegraphics[width=\linewidth]{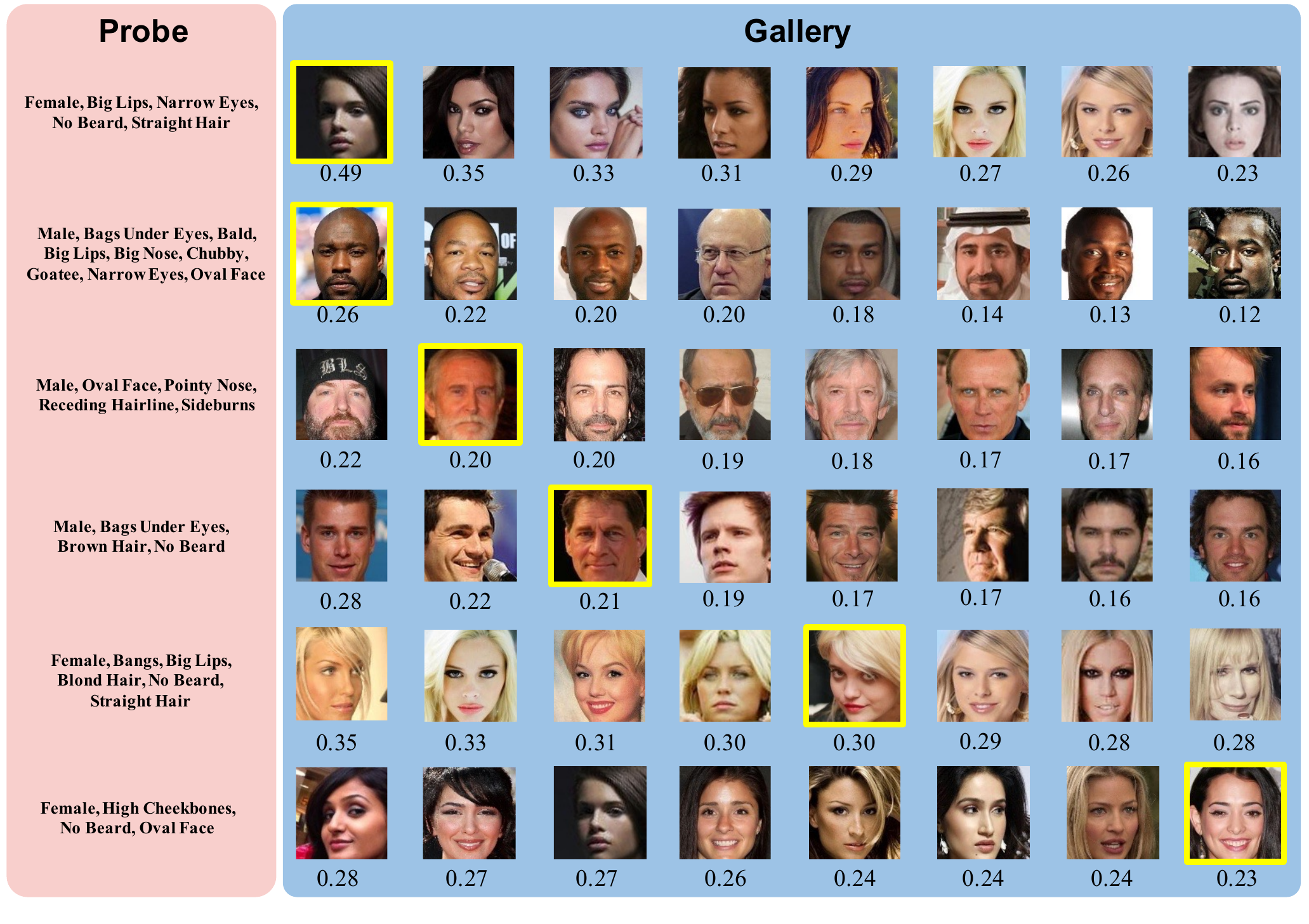}
    \caption{Visualization of face identification. Each image originates from distinct subjects. The ranking of predicted similarity decreases from left to right, accompanied by the scores presented below the images. The images enclosed within the yellow frame represent the ground truth facial image corresponding to the description provided on the left.}
    \label{fig:visualization}
\end{figure*}

In order to facilitate a comprehensive comprehension of the text-to-image face recognition process, we hereby present the visualized outcomes within this specific section. 
Demonstrating a portion of our authentic results visually becomes imperative due to the inclusion of textual language as one of the modalities in this investigation, thereby distinguishing it from the conventional face recognition approach that solely relies on image matching.
In contrast, akin to the ordinary face recognition paradigm, the objectives of the tasks at hand are further categorized into verification and identification as well.

In Fig.~\ref{fig:visualization_verification}, we illustrate the visualized results in the verification task. 
The correct predictions are depicted against a green background, whereas the incorrect ones are represented with a red background.
As an example, our model demonstrates precise negative classification for the negative pairs displayed in the four images at the bottom right corner indicated by a green background, where the DiFace takes into account various details. 
Notably, in the third image, the model identifies a discrepancy between the facial image and the accompanying descriptions such as ``pale skin" or ``wavy hair," despite partial alignment with the description, such as the attribute ``female."
Regarding the failures denoted by the red background, certain instances can be attributed to sparse descriptions, such as the first pair situated in the upper right quadrant and the second pair located in the lower left quadrant. 
In these cases, the limited information provided in the descriptions hinders the model's ability to accurately match the facial attributes, leading to erroneous predictions.
Additional failures can arise due to the presence of ambiguous descriptions or indeterminate facial features. This can be exemplified by the contradictory nature between the description of ``5 o'clock shadow" and ``no beard" in the first pair located in the lower left quadrant. 
Furthermore, in the fourth pair situated in the upper right quadrant, the occlusion of eyebrows further contributes to potential difficulties or inaccuracies.

The visualization of the identification process are shown in Fig.~\ref{fig:visualization}, wherein $6$ descriptions from the probe set have been randomly chosen, and their corresponding matched facial images of top $8$ ranks are accompanied underneath by their respective scores.
The ground truth image of each description is precisely delineated by a yellow frame, while every image pertains to a distinct subject. 
In the initial two rows, the facial images exhibiting the highest degree of similarity perfectly align with the ground truth.
Despite the possibility of an image enclosed within a yellow frame not always attaining the foremost rank, as observed in rows three to six, it is noteworthy that images showcasing higher similarity scores on their left consistently adhere to the textual description in an impeccable manner.
The reason behind this phenomenon can be attributed to the inherent imprecision of textual information when contrasted with the intricate details present in visual textures, rather than indicating any deficiency in the capabilities of DiFace. 
Evidently, a majority of the facial images depicted in Fig.~\ref{fig:visualization} conform closely to the provided descriptions, thereby serving as a demonstration to the efficacy of our model in successfully discerning and prioritizing the relevant images.

These instances highlight the complexity and potential pitfalls associated with interpreting facial attributes, which can lead to erroneous outcomes in text-to-image face recognition technology.

\section{Conclusion}
In this research endeavor, we present DiFace, an innovative solution designed to achieve text-to-image face recognition by means of a meticulously controlled diffusion process. 
This approach not only effectively addresses the challenges posed by the highly intricate cross-modal face recognition scenario but also significantly broadens the scope of application for the burgeoning diffusion models.

We commence by establishing the theoretical connections between probability density transport and the clustered recognition feature embeddings. 
By employing the text-guided diffusion model alongside our specially devised refinement module, we successfully achieve a remarkable level of accuracy in both verification and identification tasks as demonstrated through impartial experiments. 
Notwithstanding certain shortcomings in recognition, it is important to note that the observed disparities are a consequence of the fundamental limitations of text-based representations in capturing the nuanced intricacies that characterize visual imagery. 
The DiFace model, on the other hand, has demonstrated its competence by effectively filtering out and prioritizing the corresponding facial images based on the given descriptions, further affirming its robustness and reliability.
The results obtained from experiments substantiate the effectiveness and reliability of our approach, further emphasizing its potential for practical applications in the field of cross-modal face recognition.

In conclusion, as ongoing advancements persist, the unresolved aspects of face recognition through verbal description will gradually be addressed, thereby paving the way for its widespread adoption and transformative impact across diverse industries.
With researchers continuously exploring the capabilities of diffusion models and pushing the boundaries of their applications, we can anticipate witnessing their profound impact across a broader spectrum of tasks. 
These advancements will undoubtedly propel the field of artificial intelligence even further.

\section*{Acknowledgments}
This study was supported in part by National Natural Science Foundation of China (NSFC, Grant Nos. 62071292, U21B2013), Science and Technology Commission of Shanghai Municipality (STCSM, Grant Nos. 18DZ2270700).

\bibliographystyle{unsrt}  
\bibliography{DiFace-r}

\end{document}